\documentclass{article}

\PassOptionsToPackage{numbers, compress,}{natbib}
\usepackage[final]{arxiv_2017}

\usepackage[T1]{fontenc}    % use 8-bit T1 fonts
\usepackage[utf8]{inputenc} % allow utf-8 input
\usepackage{amsfonts}       % blackboard math symbols
\usepackage{booktabs}       % professional-quality tables
\usepackage{nohyperref}     % hyperlinks
\usepackage{microtype}      % microtypography
\usepackage{nicefrac}       % compact symbols for 1/2, etc.
\usepackage{url}            % simple URL typesetting

\usepackage[labelfont=bf]{caption}
\usepackage[mathscr]{eucal}
\usepackage[noend]{algpseudocode}
\usepackage[subrefformat=parens]{subcaption}
\usepackage[usenames,dvipsnames,svgnames]{xcolor}

\usepackage{algorithmicx}
\usepackage{algorithm}
\usepackage{amsfonts}
\usepackage{amsmath}
\usepackage{amssymb}
\usepackage{amsthm}
\usepackage{bm}
\usepackage{enumitem}
\usepackage{float}
\usepackage{graphicx}
\usepackage{lipsum}
\usepackage{listings}
\usepackage{placeins}
\usepackage{textcomp}
\usepackage{tikz}
\usepackage{txfonts}

\usetikzlibrary{calc}
\usetikzlibrary{decorations.pathreplacing}
\usetikzlibrary{positioning}
\usetikzlibrary{shadings}
\usetikzlibrary{shadows}
\usetikzlibrary{shapes.arrows}
\usetikzlibrary{shapes}

\newcommand*{\diff}{\mathop{}\!\mathrm{d}}

\newcommand{\abs}[1]{\left\lvert #1 \right\rvert}
\newcommand{\set}[1]{\left\{#1\right\}}

\newcommand{\mt}[1]{\mathcal{#1}}

\newcommand{\Kprior}{\mathbf{K}^{\textrm{cov}}}
\newcommand{\GP}{\mathrm{GP}}

\newcommand{\din}{d^\textrm{in}_i}
\newcommand{\dout}{d^\textrm{out}_i}

\newcommand{\djin}{d^\textrm{in}_j}
\newcommand{\djout}{d^\textrm{out}_j}

\newcommand{\dnin}{d^\textrm{in}}
\newcommand{\dnout}{d^\textrm{out}}

\newenvironment{nscenter}
 {\parskip=-2pt\par\nopagebreak\centering}
 {\par\noindent\ignorespacesafterend}

\algnewcommand{\LineComment}[1]{\State \(\triangleright\) #1}

\title{Time Series Structure Discovery\\via Probabilistic Program Synthesis}

\author{
  Ulrich Schaechtle\textsuperscript{*},
    Feras Saad\textsuperscript{*},
    Alexey Radul,
    and Vikash Mansinghka\\
  Probabilistic Computing Project\\
  Massachusetts Institute of Technology\\
  \texttt{\{ulli,fsaad,axch,vkm\}@mit.edu}
}

\begin{document}

\maketitle

\begin{abstract}
There is a widespread need for techniques that can discover structure from time
series data. Recently introduced techniques such as Automatic Bayesian
Covariance Discovery (ABCD) provide a way to find structure within a single time
series by searching through a space of covariance kernels that is generated
using a simple grammar. While ABCD can identify a broad class of temporal
patterns, it is difficult to extend and can be brittle in practice. This paper
shows how to extend ABCD by formulating it in terms of probabilistic program
synthesis. The key technical ideas are to (i) represent models using abstract
syntax trees for a domain-specific probabilistic language, and (ii) represent
the time series model prior, likelihood, and search strategy using probabilistic
programs in a sufficiently expressive language. The final probabilistic program
is written in under 70 lines of probabilistic code in Venture. The paper
demonstrates an application to time series clustering that involves a
non-parametric extension to ABCD, experiments for interpolation and
extrapolation on real-world econometric data, and improvements in accuracy over
both non-parametric and standard regression baselines.
\end{abstract}

\section{Introduction}

Time series data are widespread, but discovering structure within and among time
series can be difficult. Recent work by \citet{duvenaud2013} and
\citet{lloyd2014} showed that it is possible to learn the structure of Gaussian
Process covariance kernels and thereby discover interpretable structure in time
series data. This paper shows how to reimplement the ABCD approach from
\citet{duvenaud2013} using under 70 lines of probabilistic code in Venture
\citep{mansinghka2014}. We formulate structure discovery as a form of
``probabilistic program synthesis''. The key idea is to represent probabilistic
models using abstract syntax trees (ASTs) for a domain-specific language, and
then use probabilistic programs to specify the AST prior, model likelihood, and
search strategy over models given observed data.

Several recent projects have applied probabilistic programming techniques to
Gaussian process time series. \citet{schaechtle2015} embed GPs into Venture with
fully Bayesian learning over a limited class of covariance structures with a
heuristic prior. \citet{tong2016} describe a technique for learning GP
covariance structures using a relational variant of ABCD, and then compile the
models into Stan \citep{carpenter2017}. However, probabilistic programming is
only used for prediction, not for structure learning or for hyperparameter
inference.

The contributions of this paper are as follows. First, we formulate ABCD as
probabilistic program synthesis. Second, our implementation supports
combinations of gradient-based search for hyperparameters, and
Metropolis-Hastings sampling for structure and hyperparameters. Third, we show
competitive performance on extrapolation and interpolation tasks from real-world
data against several baselines. Fourth, we show that 10 lines of code are
sufficient to extend ABCD into a nonparametric Bayesian clustering technique
that identifies time series which share covariance structure.

\section{Bayesian structure learning as probabilistic program synthesis}
\label{sec:probabilistic-program-synthesis}

%!TEX root = ../main.tex

\begin{figure}[t]

\footnotesize

\begin{tikzpicture}[thick]

\def\bordercolor{none}

\node[
    rectangle,
    draw = black,
    text width = 2cm,
    align = center
] (ppev1){\bf
    Probabilistic\\
    Programming\\
    Engine
};

\node[
    rectangle,
    draw = \bordercolor,
    left = .3 of ppev1,
    text width = 4cm,
    align = center,
] (synthesis-inference-strategy){\bf
    Synthesis Inference Strategy
};

\node[
    rectangle,
    draw = \bordercolor,
    align = center,
    text width = 4.5cm,
    inner sep = 0cm,
    above = .75 of synthesis-inference-strategy,
] (synthesis-model) {\bf
    Synthesis Model\\(AST Prior + AST Interpreter)
};

\node[
    rectangle,
    draw = \bordercolor,
    below = .75 of synthesis-inference-strategy,
    text width = 2.5cm,
    inner sep = 0cm,
    align = center
] (dataset){
    \textbf{Observed Dataset} $\mt{D} = \set{(\din,\dout)}$
};

\node[
    rectangle,
    draw = \bordercolor,
    align = center,
    text width = 1.5cm,
    right = .5 of ppev1,
](ast-diagram) {
    \includegraphics[width=\linewidth]{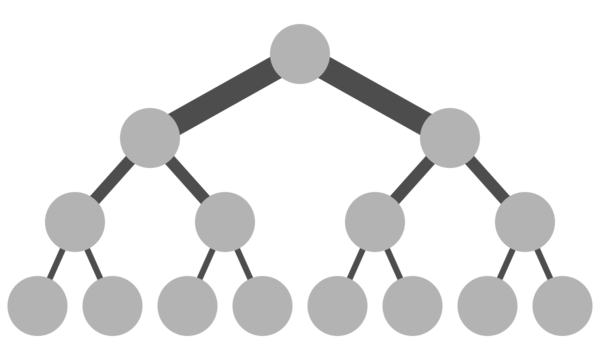}
};

\node[
    rectangle,
    draw = \bordercolor,
    below = 0 of ast-diagram.south,
    anchor = north,
    text width = 3cm,
    align = center,
    yshift = .15cm,
] (ast) {\bf\scriptsize
    Synthesized AST
};

\node[
    rectangle,
    draw = \bordercolor,
    align = center,
    text width = 3cm,
    inner sep = 0cm,
] (prediction-model) at (synthesis-model -| ast-diagram) {\bf
    Prediction Model\\(AST Interpreter)
};

\node[
    rectangle,
    draw = black,
    right = .35 of ast-diagram,
    text width = 2cm,
    align = center
] (ppev2){\bf
    Probabilistic\\
    Programming\\
    Engine
};

\node[
    rectangle,
    draw = \bordercolor,
    right = .35 of ppev2,
    text width = 2.25cm,
    align = center,
    inner sep = 0cm,
] (probe-output) {
    \textbf{Probe Outputs}\\$\set{\djout}$
};

\node[
    rectangle,
    draw = \bordercolor,
    text width = 3cm,
    align = center,
    inner sep = 0cm,
    % below = .5 of ppev2,
] (probe-input) at (dataset -| ppev2) {
    \textbf{Probe Inputs}\\$\set{\djin}$
};

\draw [-latex, thick] (synthesis-model) -- (ppev1.170);
\draw [-latex, thick] (synthesis-inference-strategy.east) -- (ppev1.180);
\draw [-latex, thick] (dataset.east) -- (ppev1.190);
\draw [-latex, thick] (ppev1.east) -- (ast-diagram);
\draw [-latex, thick] (prediction-model) -- (ppev2.170);
\draw [-latex, thick] (ast-diagram) -- (ppev2.180);
\draw [-latex, thick] (probe-input) -- (ppev2);
\draw [-latex, thick] (ppev2) -- (probe-output);
\end{tikzpicture}
\smallskip
\caption{Overview of Bayesian structure learning as probabilistic
program synthesis.}
\label{fig:synthesis-overview}

\end{figure}
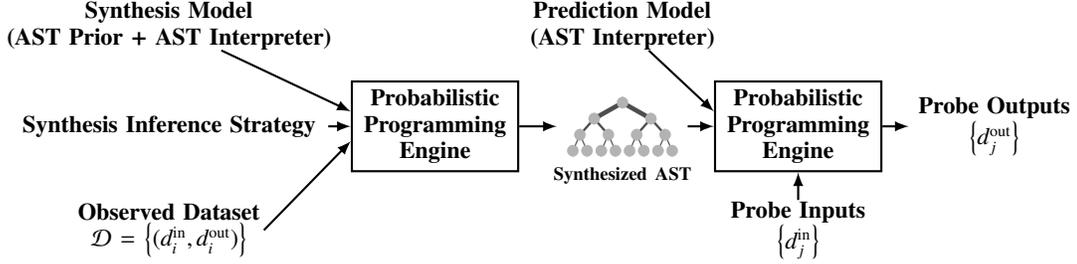

%!TEX root = ../main.tex

\begin{figure}[t]
\centering

\def\bordercolor{black}
\def\bordercolorcell{none}

\begin{tikzpicture}[thick]

%%%%%%%%%%%%%%%%%%%%%%%%%%%%%%%%%%%%%%%%%%%%%%%%%%%%%%%%%%%%%%%%%%%%%%%%%%%%%%%%
% Headers
%%%%%%%%%%%%%%%%%%%%%%%%%%%%%%%%%%%%%%%%%%%%%%%%%%%%%%%%%%%%%%%%%%%%%%%%%%%%%%%%

\node[
    draw=\bordercolorcell,
    rectangle,
    % minimum width = .3\linewidth,
    minimum height = 1.2cm,
    text width = .3\linewidth,
](header-source-tree) {\begin{nscenter}
\footnotesize\bf
synthesized abstract syntax tree of model program
\end{nscenter}
};

\node[
    draw=\bordercolorcell,
    rectangle,
    right = 0cm of header-source-tree,
    % minimum width = .31\linewidth,
    minimum height = 1.2cm,
    text width = .31\linewidth,
](header-source-code) {\begin{nscenter}
\footnotesize\bf
equivalent Venture code\\of model program
\end{nscenter}
};

\node[
    draw=\bordercolorcell,
    rectangle,
    right = 0cm of header-source-code,
    % minimum width = .35\linewidth,
    minimum height = 1.2cm,
    text width = .35\linewidth,
](header-source-exec) {\begin{nscenter}
\footnotesize\bf
Gaussian process datasets from model program executions
\end{nscenter}
};

\node[
    draw=\bordercolorcell,
    rectangle,
    below = 0cm of header-source-tree.south,
    anchor = north,
    minimum width = .33\linewidth,
    minimum height = 3.85cm,
](ast-cell-cp) {\centering\scriptsize\bf
  % cp ast cell
};

\node[
    draw=\bordercolor,
    ellipse,
    below = 0.3cm of ast-cell-cp.north,
](ast-cell-cp-node0) {\centering\scriptsize\bf
  CP 4.5 0.1
};
\node[
    draw=\bordercolor,
    ellipse,
    below left = .5cm and .1cm of ast-cell-cp-node0,
](ast-cell-cp-node1) {\centering\scriptsize\bf
  $\times$
};
\node[
    draw=\bordercolor,
    ellipse,
    below left = 1cm and 0cm of ast-cell-cp-node1,
    inner xsep = .25,
    inner ysep = 1,
](ast-cell-cp-node2) {\centering\scriptsize\bf
  LIN .36
};
\node[
    draw=\bordercolor,
    ellipse,
    below right = 1cm and 0cm of ast-cell-cp-node1,
    inner xsep = .25,
    inner ysep = 1,
](ast-cell-cp-node3) {\centering\scriptsize\bf
  WN 2.05
};
\node[
    draw=\bordercolor,
    ellipse,
    below right = .5cm and .1cm of ast-cell-cp-node0,
    inner xsep = .25,
    inner ysep = 1,
](ast-cell-cp-node4) {\centering\scriptsize\bf
  WN 11
};

\draw[-,color=\bordercolor]
  (ast-cell-cp-node0.south west) -- (ast-cell-cp-node1.north);
\draw[-,color=\bordercolor]
  (ast-cell-cp-node0.south east) -- (ast-cell-cp-node4.north);
\draw[-,color=\bordercolor]
  (ast-cell-cp-node1.south west) -- (ast-cell-cp-node2.north);
\draw[-,color=\bordercolor]
  (ast-cell-cp-node1.south east) -- (ast-cell-cp-node3.north);
% \draw[-,color=\bordercolor]
%   (ast-cell-cp-node2.south west) -- (ast-cell-cp-node5.north);
% \draw[-,color=\bordercolor]
%   (ast-cell-cp-node2.south east) -- (ast-cell-cp-node6.north);

%%%%%%%%%%%%%%%%%%%%%%%%%%%%%%%%%%%%%%%%%%%%%%%%%%%%%%%%%%%%%%%%%%%%%%%%%%%%%%%%
% Source code
%%%%%%%%%%%%%%%%%%%%%%%%%%%%%%%%%%%%%%%%%%%%%%%%%%%%%%%%%%%%%%%%%%%%%%%%%%%%%%%%

% White noise source cell.
% \node[
%     draw=\bordercolorcell,
%     rectangle,
%     below = 0cm of header-source-code.south,
%     anchor = north,
%     minimum width = .33\linewidth,
%     minimum height = 3cm,
%     text width = .33\linewidth,
%     inner sep = 0cm,
% ](source-cell-white) {
% \begin{lstlisting}
% assume covariance_kernel =
%     gp_cov_scale(
%       6.23, gp_cov_delta))
% assume gp = make_gp(
%   gp_mean_const(0.),
%   covariance_kernel)
% \end{lstlisting}
% };

% Sum source cell.
% \node[
%     draw=\bordercolorcell,
%     rectangle,
%     below = 0cm of source-cell-white,
%     anchor = north,
%     minimum width = .33\linewidth,
%     minimum height = 3cm,
%     text width = .33\linewidth,
%     inner sep = 0cm,
% ](source-cell-sum) {
% \begin{lstlisting}
% assume covariance_kernel =
%   gp_cov_sum(
%     gp_cov_periodic(.3, 8.41),
%     gp_cov_scale(
%       3.45, gp_cov_delta))
% assume gp = make_gp(
%   gp_mean_const(0.),
%   covariance_kernel)
% \end{lstlisting}
% };

% CP source cell.
\node[
    draw=\bordercolorcell,
    rectangle,
    below = 0cm of header-source-code.south,
    minimum width = .33\linewidth,
    minimum height = 3.85cm,
    text width = .33\linewidth,
    inner sep = 0cm,
    anchor = north,
](source-cell-cp) {
\begin{lstlisting}
     assume covariance_kernel =
       gp_cov_cp(
         4.5, .1,
         gp_cov_product(
           gp_cov_linear(.36),
           gp_cov_scale(
             2.05, gp_cov_delta)),
         gp_cov_scale(
           11, gp_cov_delta));
     assume gp = make_gp(
       gp_mean_const(0.),
       covariance_kernel);
\end{lstlisting}
};

%%%%%%%%%%%%%%%%%%%%%%%%%%%%%%%%%%%%%%%%%%%%%%%%%%%%%%%%%%%%%%%%%%%%%%%%%%%%%%%%
% Executions
%%%%%%%%%%%%%%%%%%%%%%%%%%%%%%%%%%%%%%%%%%%%%%%%%%%%%%%%%%%%%%%%%%%%%%%%%%%%%%%%

% White noise executions cell.
% \node[
%     draw=\bordercolorcell,
%     rectangle,
%     below = 0cm of header-source-exec.south,
%     anchor = north,
%     minimum width = .33\linewidth,
%     minimum height = 3cm,
%     text width = .33\linewidth,
%     % inner sep = 0cm,
% ](exec-cell-white) {\centering
%   \quad\;\includegraphics[width=.9\linewidth]{assets/gp-white.pdf}
% };

% Sum executions cell.
% \node[
%     draw=\bordercolorcell,
%     rectangle,
%     below = 0cm of exec-cell-white.south,
%     anchor = north,
%     minimum width = .33\linewidth,
%     minimum height = 3cm,
%     text width = .33\linewidth,
%     % inner sep = 0cm,
% ](exec-cell-sum) {\centering
%   \quad\;\includegraphics[width=.9\linewidth]{assets/gp-sum.pdf}
% };

% CP executions cell.
\node[
    draw=\bordercolorcell,
    rectangle,
    below = 0cm of header-source-exec.south,
    anchor = north,
    minimum width = .33\linewidth,
    minimum height = 3.85cm,
    text width = .33\linewidth,
](exec-cell-cp) {\centering
  \quad\;\includegraphics[width=\linewidth]{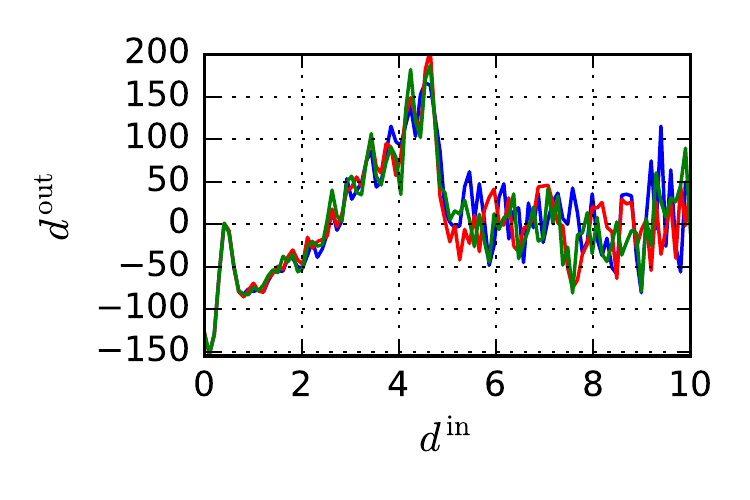}
};

%%%%%%%%%%%
%% Lines %%
%%%%%%%%%%%

\draw[line width=2pt] (header-source-tree.north west) -- (header-source-exec.north east);
\draw (header-source-tree.south west) -- (header-source-exec.south east);
% \draw (ast-cell-white.south west) -- (exec-cell-white.south east);
% \draw (ast-cell-sum.south west) -- (exec-cell-sum.south east);
% \draw (ast-cell-cp.south west) -- (exec-cell-cp.south east);
\end{tikzpicture}

\caption{Executing synthesized model programs to produce Gaussian process
datasets. \textbf{Left}: A symbolic structure generated by the AST prior.
\textbf{Center}: Equivalent source code of the Venture GP model program,
produced by the AST interpreter. \textbf{Right}: Executions of the model program
probed with inputs in the region $[0,10]$, which outputs datasets of GP time
series.}
\label{fig:execution-ast-source-series}
\end{figure}
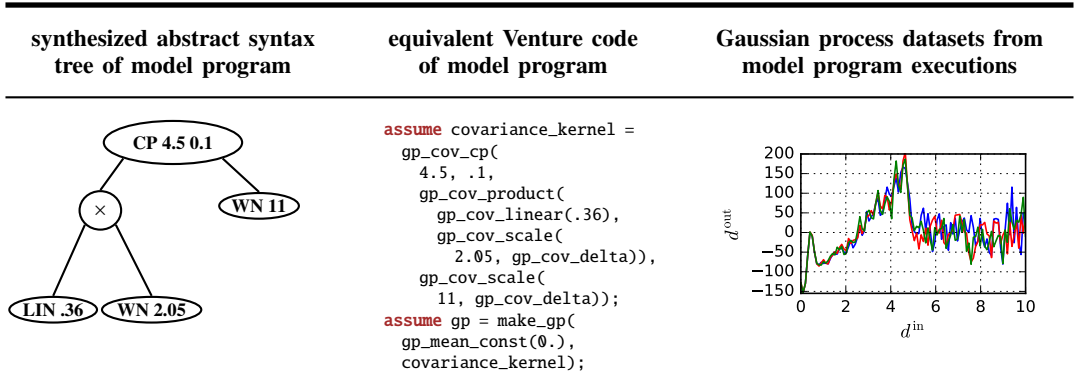

Our objective in probabilistic program synthesis for Bayesian structure learning
is to learn a symbolic representation of a probabilistic model program, by
observing outputs of the model program given the inputs at which it was
evaluated.
The basic idea is to define a joint probabilistic model over (i) the symbolic
representation of the program in terms of an abstract syntax tree (AST); (ii)
the model program synthesized from the AST; and (iii) data %
$\mathcal{D} = \{(\din, \dout): i=1,\dots,N\}$ %
that specifies constraints on observed input-output behavior of $N$ independent
executions of the synthesized model program. This framework, summarized in
Figure~\ref{fig:synthesis-overview}, is implemented using:

\begin{enumerate}

\item A pair of probabilistic programs, an AST prior and AST interpreter, which
together form the synthesis model. The AST prior $\mt{G}$ specifies a generative
model over ASTs $\mt{T}$ for a class of probabilistic model programs, denoted
$p_{\mt{G}}(\mt{T})$. The AST interpreter $\mt{I}$  takes as input $\mt{T}$ and
synthesizes an executable model program $\mt{M}$ from it. The interpreter's
distribution over model programs is $p_{\mt{I}}(\mt{M}|\mt{T})$.

\item A synthesized probabilistic model program $\mt{M}$, whose
structure and hyperparameters are determined by its AST, with a distribution
$P_\mt{M}(\dnout|\dnin)$ over output data given input data.

\item A probabilistic inference program named the synthesis strategy. Given $N$
input-output data pairs $\mt{D}$ generated by an unknown model program $\mt{M}$
from the class of programs specified by the synthesis model, it searches over
the execution trace of $\mt{G}$ to find probable symbolic structures (i.e. ASTs)
that explain the data.
\end{enumerate}

Given the description of programs above, the posterior distribution over
symbolic structures which the synthesis strategy targets is:
\begin{align}
P_{\mt{G}}(\mt{T}|\mt{D}) \propto
  p_{\mt{G}}(\mt{T})
  \int_{\mt{M}}\left(
      \prod_{i=1}^{n}P_\mt{M}(\dout|\din)
  \right) P_{\mt{I}}(\mt{M}|\mt{T})\diff{\mt{M}}.
\label{eq:full-joint-posterior}
\end{align}

% What the input and output datapoint actually represent depends on both the
% application domain as well the class of probabilistic model programs we are
% trying to learn. For example, in the time series analysis application used
% throughout this paper, we consider a class of probabilistic model programs which
% specify a sampler for a Gaussian process; the symbolic representation of the
% model programs is the covariance function of the GP, specified as a set of base
% kernels, their hyperparameters, and the operators used to compose them; the
% input data corresponds to timestamps; and the output data corresponds to
% measurements of variables at those points in time.

%!TEX root = ../main.tex

\begin{figure}[ht]
\centering

\begin{subfigure}[t]{\linewidth}
\captionsetup{skip=0pt}
\subcaption{Synthesis model: AST prior $\mt{G}$}
\label{subfig:venturescript-ast-generator}
% \hrule\smallskip
\begin{lstlisting}[frame=single, numbers=left]
assume tree_root = () -> {1};

assume get_hyper_prior ~ mem((node_index) -> {
  // Gradient-safe exponential prior.
  -log_logistic(log_odds_uniform() #hypers:node_index)
});

assume choose_primitive = (node) -> {
  base_kernel ~ categorical(simplex(.2, .2, .2, .2, .2),
    ["WN", "C", "LIN", "SE", "PER"]) #structure:pair("base_kernel", node);
  cond(
    (base_kernel == "WN")  (["WN", get_hyper_prior(pair("WN", node))]),
    (base_kernel == "C")   (["C", get_hyper_prior(pair("C", node))]),
    (base_kernel == "LIN") (["LIN", get_hyper_prior(pair("LIN", node))]),
    (base_kernel == "SE")  (["SE", .01 + get_hyper_prior(pair("SE", node))]),
    (base_kernel == "PER") (["PER",
      .01 + get_hyper_prior(pair("PER_l", node)),
      .01 + get_hyper_prior(pair("PER_t", node))
    ]))
};

assume choose_operator = mem((node) -> {
  operator_symbol ~ categorical(simplex(0.45, 0.45, 0.1),
    ["+", "*", "CP"]) #structure:pair("operator", node);
  if (operator_symbol == "CP") {
    [operator_symbol, get_hyper_prior(pair("CP", node))]
  } else {
    operator_symbol
  }
});

assume generate_random_program = mem((node) -> {
  if (flip(.3) #structure:pair("branch", node)) {
    operator ~ choose_operator(node);
    [operator, generate_random_program(2 * node), generate_random_program(2 * node + 1)]
  } else {
    choose_primitive(node)
  }
});
\end{lstlisting}
\end{subfigure}

\begin{subfigure}[t]{\linewidth}
\captionsetup{skip=0pt}
\subcaption{Synthesis model: AST interpreter $\mt{I}$}
\label{subfig:venturescript-source-synthesizer}
\begin{subfigure}{\linewidth}
\begin{lstlisting}[frame=single, numbers=left]
assume produce_covariance = (source) -> {
  cond(
    (source[0] == "WN")  (gp_cov_scale(source[1], gp_cov_bump)),
    (source[0] == "C")   (gp_cov_const(source[1])),
    (source[0] == "LIN") (gp_cov_linear(source[1])),
    (source[0] == "SE")  (gp_cov_se(source[1]**2)),
    (source[0] == "PER") (gp_cov_periodic(source[1]**2, source[2])),
    (source[0] == "+") (
      gp_cov_sum(produce_covariance(source[1]), produce_covariance(source[2]))),
    (source[0] == "*") (
      gp_cov_product(produce_covariance(source[1]), produce_covariance(source[2]))),
    (source[0][0] == "CP") (
      gp_cov_cp(source[0][1], .1, produce_covariance(source[1]), produce_covariance(source[2]))))
};

assume produce_executable = (source) -> {
  baseline_noise = gp_cov_scale(.1, gp_cov_bump);
  covariance_kernel = gp_cov_sum(produce_covariance(source), baseline_noise);
  make_gp(gp_mean_const(0.), covariance_kernel)
};
\end{lstlisting}
\end{subfigure}
\end{subfigure}

\begin{subfigure}{\linewidth}
\centering

\begin{subfigure}{.5\linewidth}
\captionsetup{skip=0pt}
\subcaption{Data observation program}
\label{subfig:venturescript-observation-program}
\begin{lstlisting}[numbers=left, frame=ltb]
assume source ~ generate_random_program(tree_root());
assume gp_executable = produce_executable(source);
define xs = get_data_xs("./data.csv");
define ys = get_data_ys("./data.csv");
observe gp_executable(${xs}) = ys;
\end{lstlisting}
\end{subfigure}%
\begin{subfigure}{.5\linewidth}
\captionsetup{skip=0pt}
\subcaption{Synthesis inference strategy: MH + Gradients}
\label{subfig:venturescript-inference-program}

\begin{lstlisting}[frame=single]
for_each(arange(T), (_) -> {
  infer gradient(
    minimal_subproblem(/?hypers), steps=100);
  infer resimulate(
    minimal_subproblem(one(/?structure)), steps=100)})
\end{lstlisting}
\end{subfigure}%
\end{subfigure}

\caption{Structure discovery in time series via probabilistic program
synthesis.  Panels \subref{subfig:venturescript-ast-generator}--\subref{subfig:venturescript-inference-program}
present interacting programs in Venture which
correspond to the components in Figure~\ref{fig:synthesis-overview}.}
\label{fig:venturescript}
\end{figure}

\section{Applying the framework to Bayesian learning of Gaussian process
covariance structures}
\label{sec:gaussian-process}

Recent work by \citet{duvenaud2013} and \citet{lloyd2014} showed it is possible
to use Gaussian Processes (GPs) to discover covariance structure in univariate
time series. In this section, we extend the basic approach from
\citet{duvenaud2013} by using probabilistic program synthesis for Bayesian
learning over the symbolic structure of GP covariance kernels. The technique is
implemented in under 70 lines of Venture code, shown in
Figure~\ref{fig:venturescript}.

We briefly review the Gaussian process, a nonparametric regression technique
that learns a function $f: \mathcal{X} \to \mathcal{Y}$. The GP prior can
express both simple parametric forms (such as polynomial functions) as well as
more complex relationships dictated by periodicity, smoothness, and so on.
Following the notation of \cite{rasmussen2006}, we formalize a Gaussian
process $f \sim \GP(m, k)$ with mean function $m: \mathcal{X} \to \mathcal{Y}$
and covariance function $k: \mathcal{X} \times \mathcal{X} \to \mathbb{R}$ as follows: $f$ is a
collection of random variables $\set{f(x): x \in \mathcal{X}}$, any finite
subcollection $[f(x_1), \dots, f(x_n)]$ of which are jointly Gaussian with mean
vector $[m(x_1),\dots,m(x_n)]$ and covariance matrix
$[k(x_i,x_j)]_{1\le{i,j}\le{n}}$. The mean $m$ is typically set
to zero as it can be absorbed by the covariance.
The functional form of the covariance $k$ defines essential features of the unknown function $f$ and so provides
the inductive bias which lets the GP (i) fit patterns in data, and (ii) generalize to out-of-sample
predictions. Rich GP kernels can be created by composing simple (base) kernels
through sum, product, and change-point operators %
\citep[Section 4.2.4]{rasmussen2006}.

We now describe the synthesis model (AST prior and AST interpreter), and the
class of synthesized model programs for learning GP covariances structures. The
AST prior $\mt{G}$ (Codebox~\ref{subfig:venturescript-ast-generator}) specifies
a prior over binary trees. Each leaf $n$ of $\mt{T}$ is a pair $(k_n, h_n)$
comprised of a base kernel and its hyperparmaters. The base kernels are: white
noise (WN), constant (C), linear (LIN), squared exponential (SE), and periodic
(PER). Each base kernels has a set of hyperparameters; for instance, PER has a
lengthscale and period, and LIN has an x-intercept. Each internal node $n$
represents a composition operator $o_n$, which are: sum ($+$), product
($\times$), and changepoint ($\mathrm{CP}$, whose hyperparameters are the
x-location of the change, and decay rate). The structure of $\mathcal{T}$ is
encoded by the index $n$ of each internal node (whose left child is $2n$ and
right child is $2n+1$) and the operators and base kernels at each node.
Let $N = \lvert\mt{T}\rvert$ denote the number of nodes. We write $\mt{T} =
\cup_{n=1}^{N}{\set{x_n}}$ as a collection of $N$ random variables, where $x_n =
(b_n, o_n, k_n, h_n)$ is a bundle of random variables for node $n$: $b_n$ is 1
if the tree branches at $n$ (and 0 if $n$ is a leaf); $o_n$ is the operator (or
$\varnothing$ if $b_n=0$); $k_n$ is the base kernel (or $\varnothing$ if
$b_n=1$); and $h_n$ is the hyperparameter vector (or $\varnothing$ if $n$ has no
hyperparameters, e.g., if $b_n = 1$ and $o_n = +$). Letting $\pi(n)$ denote the
list of all nodes in the path from $n$ up to the root, the tree prior is:
\begin{align}
p_{\mt{G}}(\mt{T})
  &= \prod_{n=1}^{N}p_{\mt{G}}(x_n|x_{\pi(n)});
  \label{eq:g-tree-prior} \\
% p_{\mt{G}}(x_n|x_{\pi(n)})
  &= \prod_{n=1}^{N} \begin{cases}
  (1{-}p_\mathrm{branch})\,
    p_\mathrm{kernel}(k_n)\,
    p_\mathrm{hyper}(h_n\,{\mid}\,\mathrm{kernel} = k_n)
    & \textrm{if}\; b_n = 0,\\
  (p_\mathrm{branch})
    p_\mathrm{operator}(o_n)
    p_\mathrm{hyper}(h_n\,{\mid}\,\mathrm{operator} = o_n)
  & \textrm{if}\; b_n = 1,\\
  0
  & \textrm{if}\; x_n|x_{\pi(n)}\; \textrm{is inconsistent.}
  \notag
\end{cases}
\end{align}
The distributions $p_\textrm{branch}$, $p_\textrm{kernel}$, and
$p_\mathrm{hyper}$ are all fixed constants in $\mt{G}$. An example covariance
kernel AST generated by Eq~\eqref{eq:g-tree-prior} is shown the first column of
Figure~\ref{fig:execution-ast-source-series}. As for the AST interpreter
$\mt{I}$ (Codebox~\ref{subfig:venturescript-source-synthesizer}), it parses
$\mt{T}$ and deterministically outputs a GP model program with mean 0 and
covariance function encoded by $\mt{T}$, plus baseline noise. Outcomes of the
synthesis step are shown in the second column of
Figure~\ref{fig:execution-ast-source-series}. The synthesized GP model program
$\mt{M}$ takes as input $k$ probe points $\dnin \in \mathbb{R}^k$, and produces
as output a (noisy) joint sample $\dnout \in \mathbb{R}^k$ of the GP at the probe
points:
\begin{align*}
\log P_\mt{M}(\dnout|\dnin)
&= \log \mathcal{N}(\dnout\mid 0, \Kprior +
  \sigma^2 \mathbf{I}) \quad\quad
  (
    \mathrm{with}\; \Kprior = \left[
      \mathrm{cov}\left(d^\text{in}_{a}, d^\text{in}_{b}\right)
    \right]_{1\le a,b \le k}
  )
\\
&= -\frac12 (\dnout)^\top (\Kprior + \sigma^2
  \mathbf{I})^{-1}\dnout
  - \frac12\log \abs{\Kprior + \sigma^2 \mathbf{I}}
  - \frac{k}{2}\log 2\pi.\\
\end{align*}

\FloatBarrier

\section{Bayesian structure learning and hyperparameter inference in the
covariance kernel AST}
\label{sec:bayesian-structure-learning}

Our implementation of program synthesis for GP covariances described in the
previous section simplifies Eq~\eqref{eq:full-joint-posterior} in that $\mt{I}$
deterministically interpreters a GP model program given the AST, so that
$P_{\mt{I}}(\mt{M}|\mt{T}) = \delta\left[\mt{M} = \texttt{interpret}(\mt{T};
\mt{I}) \right]$. The key inference problem becomes search over the space of GP
kernel compositions in $\mt{T}$, and hyperparameters $h_n \in \mt{T}$ of base
kernels. This section describes the synthesis strategy for posterior inference
over the AST.

Our strategy for inference on structure is to simulate a Markov chain whose
target distribution is $p_{\mt{G}}(\mt{T}|\mt{D})$. The following
Metropolis-Hastings algorithm is implemented by the Venture inference program %
\lstinline{infer resimulate(minimal_subproblem(/?structure))} (invoked in
Codebox~\ref{subfig:venturescript-inference-program}).
Suppose the current AST is $\mt{T}$. We design a proposal distribution $Q(\mt{T}
\to \mt{T}')$ using a three-step process. First, identify a node $x_n \in
\mt{T}$, and let $\mt{T}_n$ denote the subtree of $\mt{T}$ rooted at $n$. Second,
``detach'' $x_n$ and all its descendants from $\mt{T}$, which gives an
intermediate tree $\mt{T}_{\textrm{detach}} := \mt{T} \setminus \mt{T}_n$.
Third, ``resimulate'', starting from $\mt{T}_{\textrm{detach}}$, the random
choice $x'_n$ using a resimulation distribution
$q(x'_n|\mt{T}_{\textrm{detach}}) \equiv p_{\mt{G}}$ equal to the prior
Eq~\eqref{eq:g-tree-prior}. If $b_n' = 0$ (i.e., a branch node), recursively
resimulate its children until all downstream random choices are leaf nodes. This
operation results in a new subtree $\mt{T}'_{n}$, and we set $\mt{T}' =
\mt{T}_{\textrm{detach}} \cup \mt{T}'_{n}$ to be the proposal.
To compute the reversal $Q(\mt{T}'\to\mt{T})$, we make the following key
observation: when resimulating node $x'_n$ starting from $\mt{T}'$, the
intermediate trees $\mt{T}_{\mathrm{detach}} =
\mt{T}'_{\mathrm{detach}}$ are identical for $Q(\mt{T}\to\mt{T}')$ and
$Q(\mt{T}'\to\mt{T})$. Using this insight, the MH ratio is therefore:
\begin{align*}
&\alpha(\mt{T} \to \mt{T}')
= \frac{
    p_{\mt{G}}(\mt{T}', \mt{D}) Q(\mt{T}'\to\mt{T})}{
    p_{\mt{G}}(\mt{T}, \mt{D}) Q(\mt{T}\to\mt{T}')
  }
= \frac{
    \left(\prod_{n=1}^{N}p_{\mt{G}}(x'_n|x'_{\pi'(n)})\right)
    p_{\mt{G}}(\mt{D}|\mt{T}')
    Q(\mt{T}'\to\mt{T})
    }{
    \left(\prod_{n=1}^{N}p_{\mt{G}}(x_n|x_{\pi(n)})\right)
    p_{\mt{G}}(\mt{D}|\mt{T}) Q(\mt{T}\to\mt{T}')
  }\\
&= \frac{
    \left(
      \prod_{n\in{\mt{T}'_{\mathrm{detach}}}}p_{\mt{G}}(x'_n|x'_{\pi'(n)})
    \right)
    \left(
      \prod_{n\in{\mt{T}'_n}}p_{\mt{G}}(x'_n|x'_{\pi'(n)})
    \right)
      p_{\mt{G}}(\mt{D}|\mt{T}')
    \left(
      \prod_{n\in{\mt{T}_n}}p_{\mt{G}}(x_n|x_{\pi'(n)})
    \right)
    }{
    \left(
      \prod_{n\in{\mt{T}_{\mathrm{detach}}}}p_{\mt{G}}(x_n|x_{\pi(n)})
    \right)
    \left(
      \prod_{n\in{\mt{T}_n}}p_{\mt{G}}(x_n|x_{\pi(n)})
    \right)
      p_{\mt{G}}(\mt{D}|\mt{T})
    \left(
      \prod_{n\in{\mt{T}'_n}}p_{\mt{G}}(x'_n|x_{\pi'(n)})
    \right)
  }\notag
= \frac{p_{\mt{G}}(\mt{D})\mid\mt{T}')}{p_{\mt{G}}(\mt{D}\mid\mt{T})}.
\end{align*}

This likelihood-ratio can be computed without revisiting the entire trace
\citep{mansinghka2014}. Algorithm~\ref{alg:resimulation-mh} summarizes the key
elements of the MH resimulation algorithm described above.

As for hyperparameter inference, Our synthesis strategy uses either MH (for each
hyperparameter separately) or gradient ascent (for all hyperparameters jointly).
The gradient optimizer uses reverse-mode auto-differentiation
\citep{griewank2008}, propagating gradients down the root of $\mt{T}$ to the
leaves, and partial derivatives of hyperparameters from leaves back up to the
root. Algorithms~\ref{alg:hypers-mh} and \ref{alg:hypers-gradient} describe
hyperparameter inference in the AST using MH and gradient-based inference,
respectively. All three algorithms are implemented as general purpose inference
machinery in Venture.
\begin{algorithm}[H]
\small
\caption{Resimulation MH for the covariance AST.\\
Inference program: \lstinline{infer resimulate(minimal_subproblem(/?structure==pair("branch", n)))}}
\label{alg:resimulation-mh}
\begin{algorithmic}[1]
\Require{Index $n$ of node in the AST whose
  subtree structure $\mt{T}_n$ to transition.
}
\Ensure{MH transition $\mt{T}_n\to\mt{T}_n'$,
  targeting $p_{\mt{G}}(\mt{T}_n|\mt{T}_{\setminus{n}}, \mt{D})$.
}
\State $\mt{T}_{\textrm{detach}} \gets \mt{T} \setminus \mt{T}_n$
  \Comment{Detach the subtree rooted at $n$.}
\State $\mt{T}'_n \sim p_{\mt{G}}(\cdot \mid \mt{T}_{\textrm{detach}})$
  \Comment{Resimulate the subtree rooted at $n$ from the prior.}
\State $\mt{T}' \gets \mt{T}_{\textrm{detach}} \cup \mt{T}'_n$
  \Comment{Construct the proposal tree.}
\State $\alpha \gets
  p_{\mt{G}}(\mt{D}\mid\mt{T}')/p_{\mt{G}}(\mt{D}\mid\mt{T}')$
  \Comment{Compute the acceptance ratio.}
\If{$\textsc{Uniform}[0,1] \le \alpha$}
  \Comment{Accept the proposal with probability $\alpha$.}
  \State $\mt{T}_n \gets \mt{T}'_n$
\EndIf
\end{algorithmic}
\end{algorithm}\vspace{-0.75cm}
\begin{algorithm}[H]
\small
\caption{MH transition on hyperparameters of covariance base kernels and operators.\\
Inference program: \lstinline{infer resimulate(minimal_subproblem(/?hypers==n), steps=T)}}
\label{alg:hypers-mh}
\begin{algorithmic}[1]
\Require{
  Index $n$ of node in AST whose hypers to transition;
  number $T$ of MH steps.
}
\Ensure{MH transition targeting
  $p_{\mt{G}}(h_n | \mt{D}, \mt{T}\setminus{\set{h_n}})$.}
\For{$t=1,\dots, T$}
\State $h'_n \sim q(\cdot | \mt{T}, \mt{D})$
  \Comment{Propose a new value of $h'_n$.}
\State $q_{\textrm{reverse}} =
  p(\mt{T}\setminus{\set{h_n}}, h'_n, \mt{D})
  \, q(h_n | (\mt{T}\cup{h'_n})\setminus{\set{h_n}}, \mt{D})$
\Comment{Compute density of reversal proposal.}
\State $q_{\textrm{forward}} =
  p(\mt{T},\mt{D})
  \, q(h'_n | \mt{T}, \mt{D})$
\Comment{Compute density of forward proposal.}
\State $\alpha \gets q_{\textrm{reverse}} / q_{\textrm{forward}}$
  \Comment{Compute the acceptance ratio.}
\If{$\textsc{Uniform}[0,1] \le \alpha$}
  \Comment{Accept the proposal with probability $\alpha$.}
  \State $\mt{T} \gets (\mt{T}\cup{h'_n})\setminus{\set{h_n}}$
\EndIf
\EndFor
\end{algorithmic}
\end{algorithm}\vspace{-0.75cm}%
\begin{algorithm}[H]
\small
\caption{Reverse auto-differentiation jointly optimizing
  hyperparameters of all kernels.\\
Inference program: \lstinline{infer gradient(minimal_subproblem(/?hypers), steps=T, step_size=g)}}
\label{alg:hypers-gradient}
\begin{algorithmic}[1]
\algrenewcommand\algorithmicindent{.75em}
\Require{%
  Number $T$ of gradient steps; gradient step size $\gamma$.
}
\Ensure{%
  Gradient ascent on all hypers
$\bm{h} = (h_n\,{:}\,n\,{\in}\,\mt{T})$ optimizing un-normalized posterior
  $p(\mt{D}, \bm{h}, \mt{T}\setminus\set{\bm{h}})$.}
\LineComment{
Posterior factors as $
  p(\mt{D}, \bm{h}, \mt{T}\setminus\set{\bm{h}}) =
    \mt{L}(\bm{h}) p(\bm{h} | \mt{T}\setminus\set{\bm{h}})
    p(\mt{T}\setminus\set{\bm{h}}),
  $
  where $\mt{L}(\bm{h}) = p(\mt{D}|\bm{h}, \mt{T}\setminus\set{\bm{h}})$.
}
\For{$t=1,\dots,T$}
\State $(C^t_1,\dots,C^t_n) \gets \textsc{Compute-Covariance-Matrices}(\mt{T})$
  \Comment{Recompute covariance matrices at all subtrees.}
\State $\diff{C}^t_1 = \vec{\nabla} \log p(\mt{D} | C^t_1)$
  \Comment{Compute gradient of $\log\mt{L}$ wrt $C^t_1$.}
\State \textsc{Backpropagate-Gradient-Subtree}$(\mt{T}, 1)$
  \Comment{Compute gradients of $\log\mt{L}$ wrt $C^t_n$, $h_n$ at all subtrees.}
\State $\diff{\bm{h}} \gets \diff{\bm{h}}
    + \vec{\nabla} \log p(\bm{h}|\mt{T}\setminus\set{\bm{h}})$
  \Comment{Add gradient of the prior on $\bm{h}$.}
\State $\bm{h} \gets \bm{h} + \gamma \diff{\bm{h}}$
  \Comment{Jointly update all hyperparameters.}
\EndFor
\newline
\Require{
  AST $\mt{T}$;
  node index $n$;
  covariance matrix $C^t_n$
  and hyperparameters $h_n$ at $\mt{T}_n$, and at all subtrees}
\Ensure{
  Store: gradients $\diff{C}^t_j$ and hyperparam partial derivatives
  $\diff{h} = \vec{\nabla}_{h_{j,i}}(\log\mt{L}(\bm{h}))$ of all children $j$ of $n$.}
\Procedure{Backpropagate-Gradient-Subtree}{AST $\mt{T}$, node index $n$}
  \If{$b_n\; \mathrm{\bf is}\; 1$} \Comment{A branch node.}
    \If{$o_n\; \mathrm{\bf is}\; '+'$}
      \Comment{Child gradient is the parent gradient.}
      \State $\diff{C}^t_{2n} \gets \diff{C}^t_{n}$
        \Comment{Update left child.}
      \State $\diff{C}^t_{2n+1} \gets \diff{C}^t_{n}$
        \Comment{Update right child.}
    \Else{\;$(o_n\; \mathrm{\bf is}\; '\times')$}
      \Comment{
        Child gradient is element-wise product of parent gradient
        and sibling covariance.
      }
      \State $\diff{C}^t_{2n} \gets \diff{C}^t_{n} \odot C^t_{2n+1}$
        \Comment{Update left child.}
      \State $\diff{C}^t_{2n+1} \gets \diff{C}^t_{n} \odot C^t_{2n}$
        \Comment{Update right child.}
    \EndIf
  \State \textsc{Backpropagate-Gradient-Subtree}$(\mt{T}, 2n)$
    \Comment{Backpropagate down left subtree.}
  \State \textsc{Backpropagate-Gradient-Subtree}$(\mt{T}, 2n+1)$
    \Comment{Backpropagate down right subtree.}
  \Else{\;$(b_n\; \mathrm{\bf is}\; 0)$}
    \Comment{A leaf node.}
    \State $\diff{h}_n \gets \mathrm{\bf vec}(\diff{C}^t_n) \cdot \mt{J}_{h_n}
      (K_n(\mt{D}^{\rm{in}},\mt{D}^{\rm{in}}\mid{h_n}))$
    \Comment{
      Gradient is product of covariance with kernel's Jacobian.}
  \EndIf
\EndProcedure
\end{algorithmic}
\end{algorithm}

\section{Applications to synthetic and real-world datasets}
\label{sec:applications}

%!TEX root = ../main.tex

\begin{figure}[ht]

\begin{subfigure}[b]{.45\linewidth}
\includegraphics[width=\linewidth]{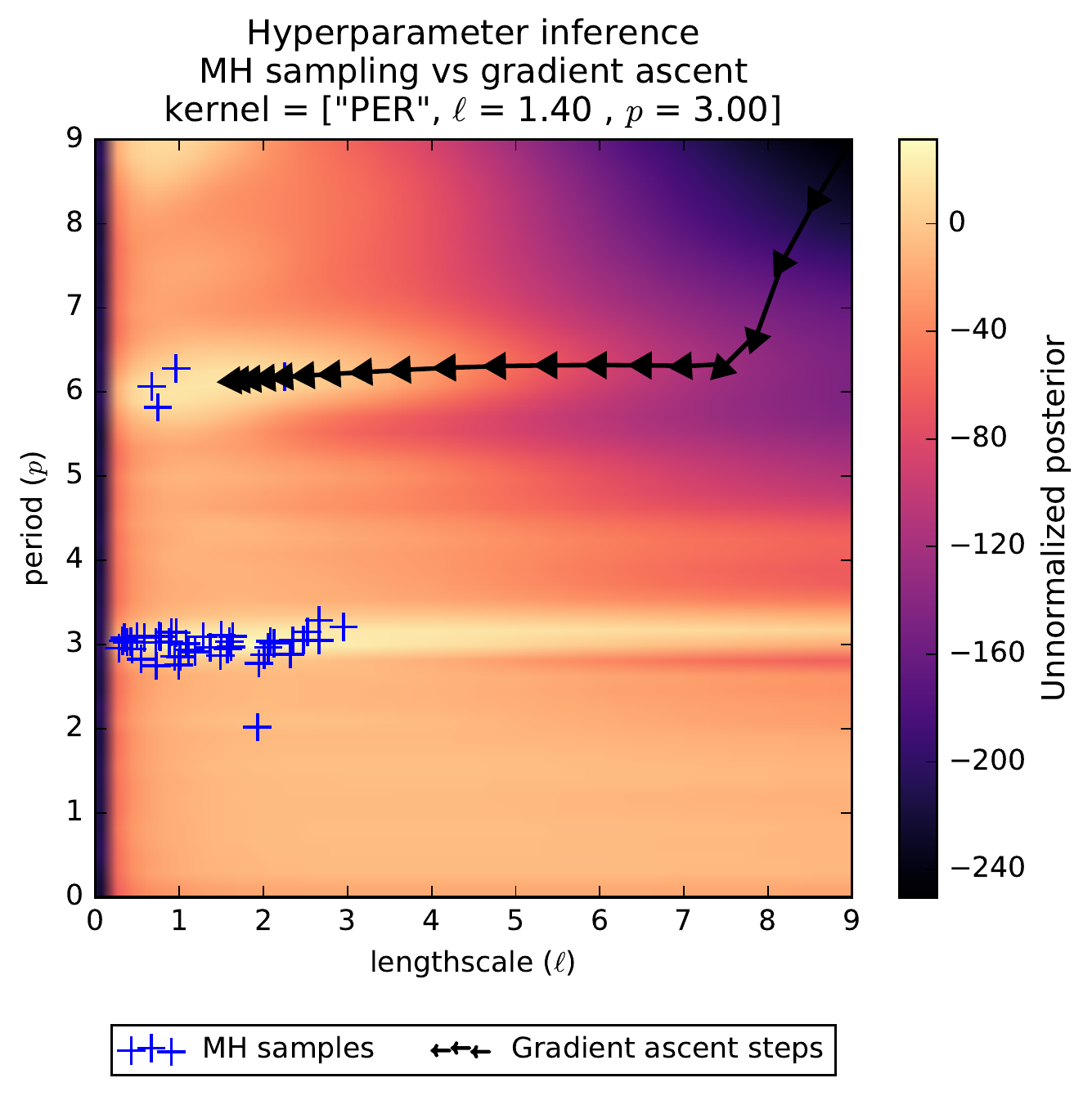}
\end{subfigure}%
\begin{subfigure}[b]{.55\linewidth}
\lstinline{infer resimulate(minimal_subproblem(one(/?hypers)), steps=100)}
\includegraphics[width=\linewidth]{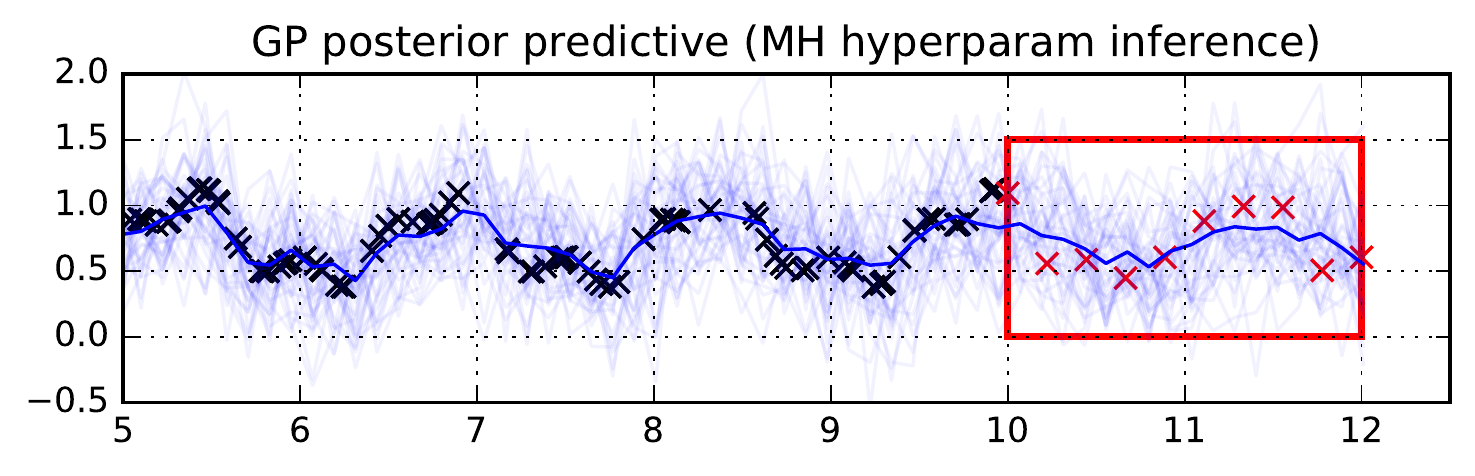}
\lstinline{infer gradient(minimal_subproblem(/?hypers), steps=100)}
\includegraphics[width=\linewidth]{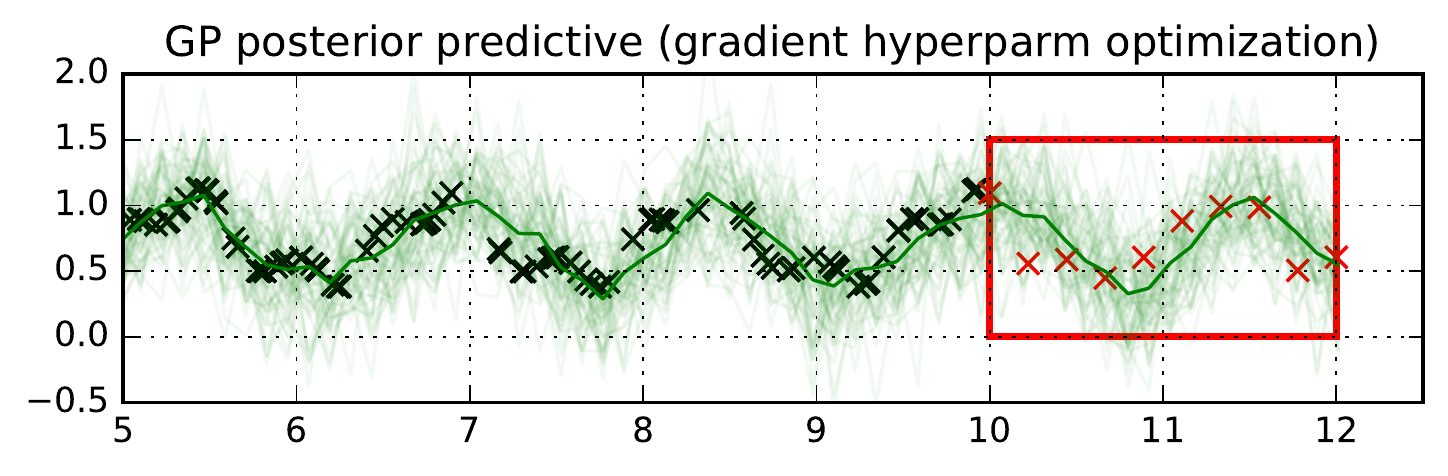}
\end{subfigure}

\caption{Comparing Metropolis-Hastings sampling and gradient ascent
as synthesis strategies for learning GP hyperparameters. The data are drawn
from a periodic kernel with length scale 1.4 and period 3.
\textbf{Left}: The surface of the
unnormalized posterior after observing 200 data points. The gradient ascent
steps converge to a local mode at period 6 (a multiple of the true period); the
MH sampler explores both posterior modes at periods of 3 and 6.
\textbf{Right}: By averaging over modes, posterior
predictive curves from the MH inference program (top) illustrate smoother
behavior than predictive curves from gradient inference (bottom), which center
on a single mode. The mean squared prediction errors on held-out data (red
crosses) are 0.18 and 0.20 for MH and gradients, respectively.}
\label{fig:hypers}

\medskip
\hrule
\smallskip

\begin{subfigure}{.5\linewidth}
\centering
\captionsetup{skip=0pt}
\includegraphics[width=\linewidth]{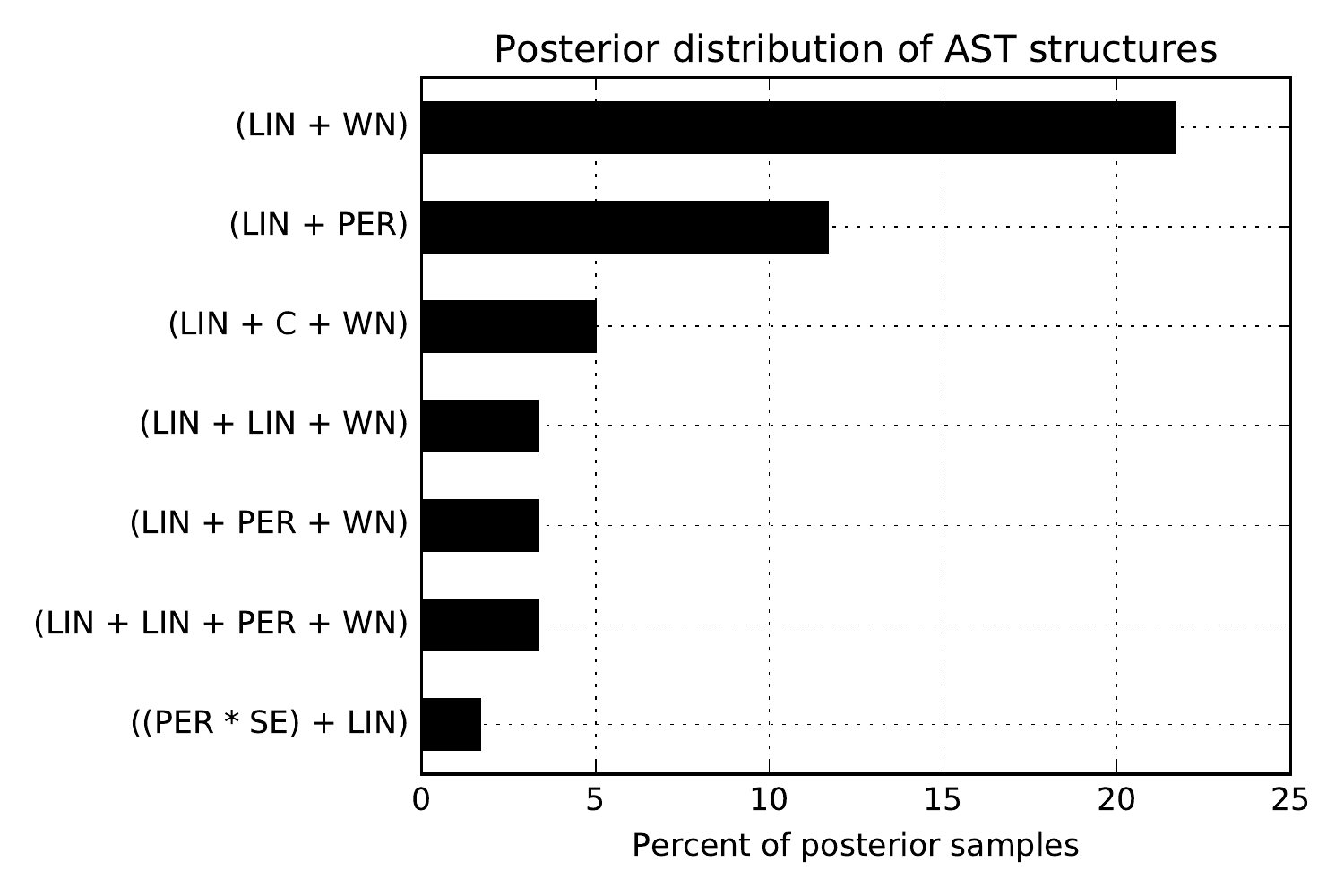}%
% \subcaption{}
\end{subfigure}%
\begin{subfigure}{.5\linewidth}
\includegraphics[width=\linewidth]{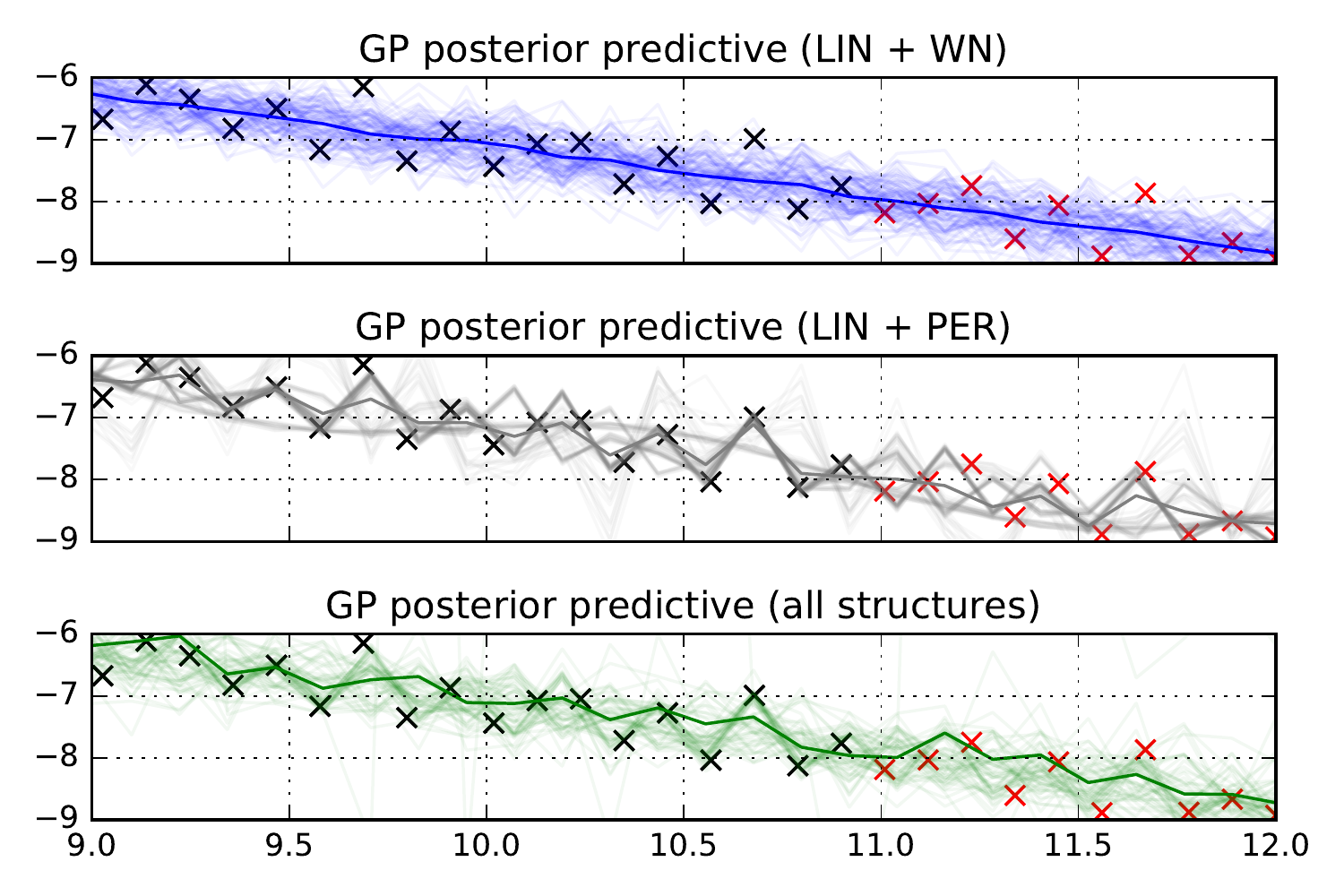}
\captionsetup{skip=0pt}
% \subcaption{}
\end{subfigure}

\caption{Structure recovery from synthetic data.
\textbf{Left}: Histogram of the posterior distribution over AST structures,
given data from a LIN + PER Gaussian process. The MAP structure (posterior mode)
is LIN + WN.
\textbf{Right}: Posterior
GP curves sampled from the LIN + WN (blue) and LIN + PER (gray) structures;
LIN + PER, which is not the MAP, achieves better predictions on the held-out
data (red crosses). Model averaging (green) smooths predictions over all
structures.}
\label{fig:structures}

\medskip
\hrule
\smallskip

\begin{subfigure}[b]{.47\linewidth}
\centering
\begin{lstlisting}
assume crp = make_crp(0.5);
assume get_cluster_id = mem((ts_index) -> {crp()});
assume get_source = (ts_index) -> {
    cluster_id ~ get_cluster_id(integer(ts_index));
    generate_random_program(
        get_tree_root(), cluster_id)
};
assume obs_function = (ts_index, data) -> {
    source  = get_source(ts_index);
    produce_executable(source)(data)
};
\end{lstlisting}
\captionsetup{skip=0pt}
\caption{Observation program for clustering GP curves.}
\label{subfig:clusters-code}
\end{subfigure} \hfill%
\begin{subfigure}[b]{.47\linewidth}
% \centering
\includegraphics[width=\linewidth]{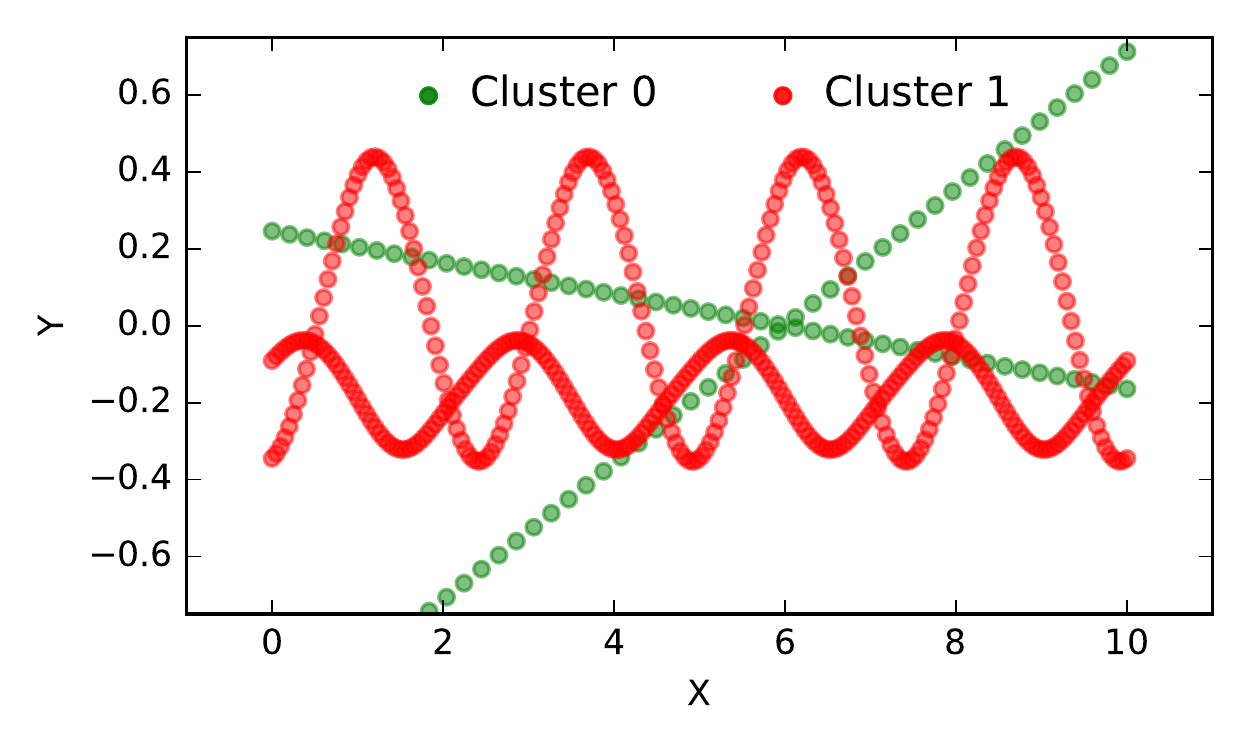}
\captionsetup{skip=0pt}
\subcaption{Observed points colored by inferred cluster.}
\label{subfig:clusters-scatter}
\end{subfigure}%

\caption{Clustering data according to structural characteristics.
\textbf{Left}: A small modification to the model program from
Codebox~\ref{subfig:venturescript-ast-generator} suffices to extend it for
clustering time series by their covariance structures. \textbf{Right}:
Detected clustering among four synthetic time series of 100 points each.}
\label{fig:clusters}

\end{figure}

% Our experiments demonstrate %
% (i) comparisons of the outcomes of different synthesis strategies, namely
% MH-sampling versus gradient-ascent for hyper-parameter learning; %
% (ii) advantages of Bayesian model averaging over the posterior distribution of
% covariance structures versus MAP estimation; %
% (iii) interpolation and extrapolation of held-out data with composite covariance
% structures; %
% (iv) a non-parametric Bayesian extension to the basic model for clustering GP
% curves; %
% (v) improvements in predictive accuracy of GP program synthesis over several
% non-parametric and standard ML regression techniques on real-world datasets; %
% and %
% (vi) reductions in lines-of-code using probabilistic program synthesis compared
% to ABCD.

In this section, we apply the probabilistic program synthesis framework for
learning GP covariance structures to a collection of synthetic and real-world
examples.

The first experiment compares the outcomes of hyperparameter inference using two
different inference programs: MH sampling (Algorithm~\ref{alg:hypers-mh}) and
gradient optimization (Algorithm~\ref{alg:hypers-gradient}), given data from a
periodic GP with period 3 and lengthscale 1.4. By encapsulating inference
algorithms as top-level inference programs, it is possible to easily compare
both their performance in searching the hyperparameter space, and their
predictive outcomes. Refer to the figure caption for further details.

To explore the advantage of Bayesian learning versus greedy search over
structures, we ran inference on 50 data points from a GP with a LIN + PER
composite covariance kernel. The posterior distribution over structures is shown
in Figure~\ref{fig:structures}. The ground truth structure is the second most
probable, while the MAP estimate is incorrectly identified as LIN + WN. GP
predictives from model averaging  over the posterior structure distribution
provide a better fit than using the MAP structure.

To assess the flexibility and extensibility of time series discovery as
probabilistic program synthesis, we extended the observation program from
Codebox~\ref{subfig:venturescript-observation-program} to specify a
non-parametric mixture of several GP curves, as shown in
Figure~\ref{subfig:clusters-code}. We simulated four datasets, (two linear, and
two periodic), and then ran joint MH inference over their structures,
hyperparameters, and cluster identities. Clusterings based on 64 posterior
samples correctly recover the ground-truth partitioning, shown in red and green
in Figure~\ref{subfig:clusters-scatter}. It is worthwhile to note that this
significant change to the probabilistic model is achieved by modifying less than
10 lines of the original code, suggesting it is possible to extend the basic
synthesis template from Figure~\ref{fig:venturescript} to a variety of time
series analysis tasks.

\begin{figure}[ht]

%% Figure: Interpolation and Extrapolation.
\begin{subfigure}{\linewidth}
\begin{subfigure}{.5\linewidth}
\includegraphics[width=\linewidth]{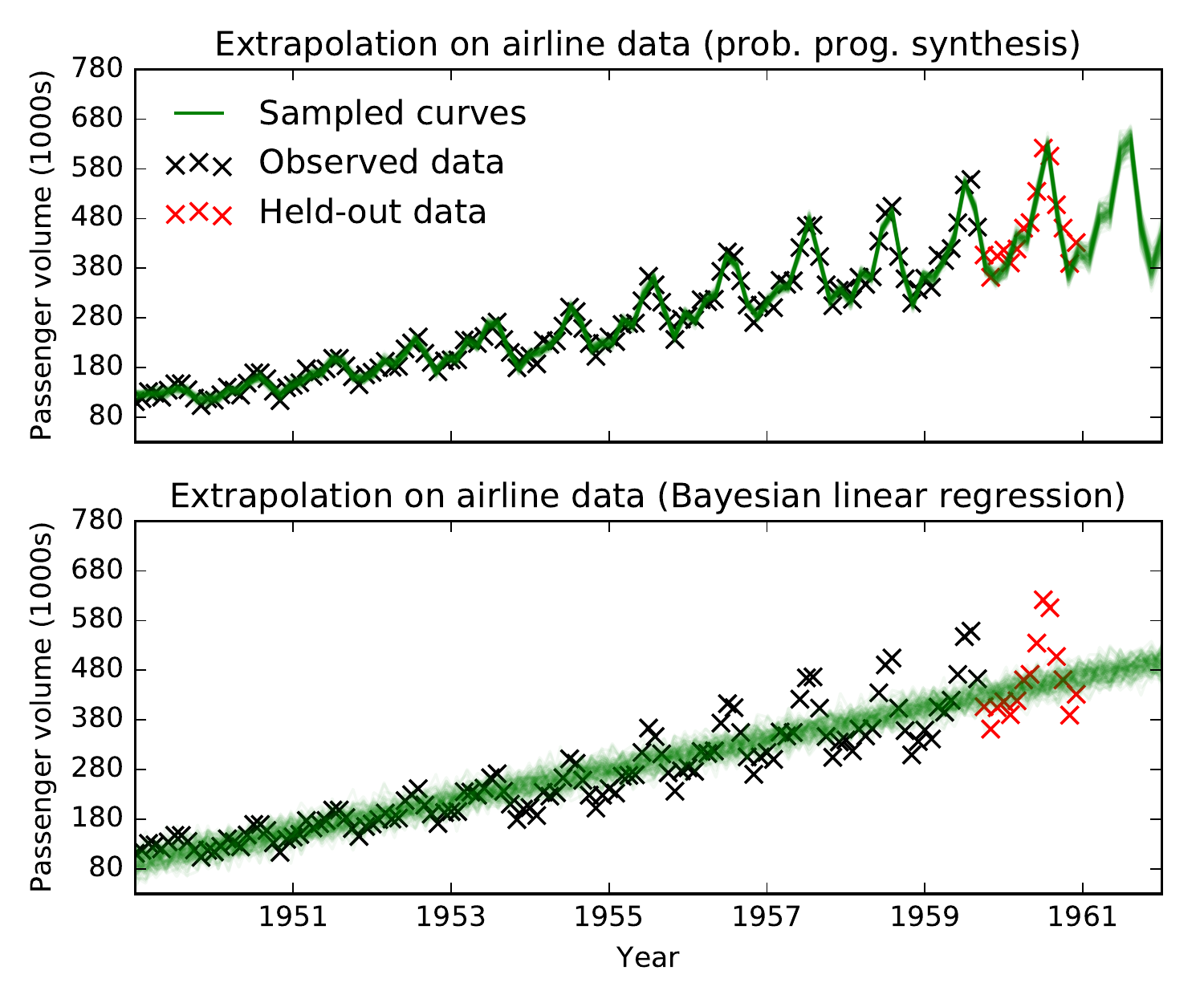}
\captionsetup{skip=0pt}
\subcaption{Extrapolation of airline travel volume}
\label{subfig:predictions-extrapolation}
\end{subfigure}%
\begin{subfigure}{.5\linewidth}
\includegraphics[width=\linewidth]{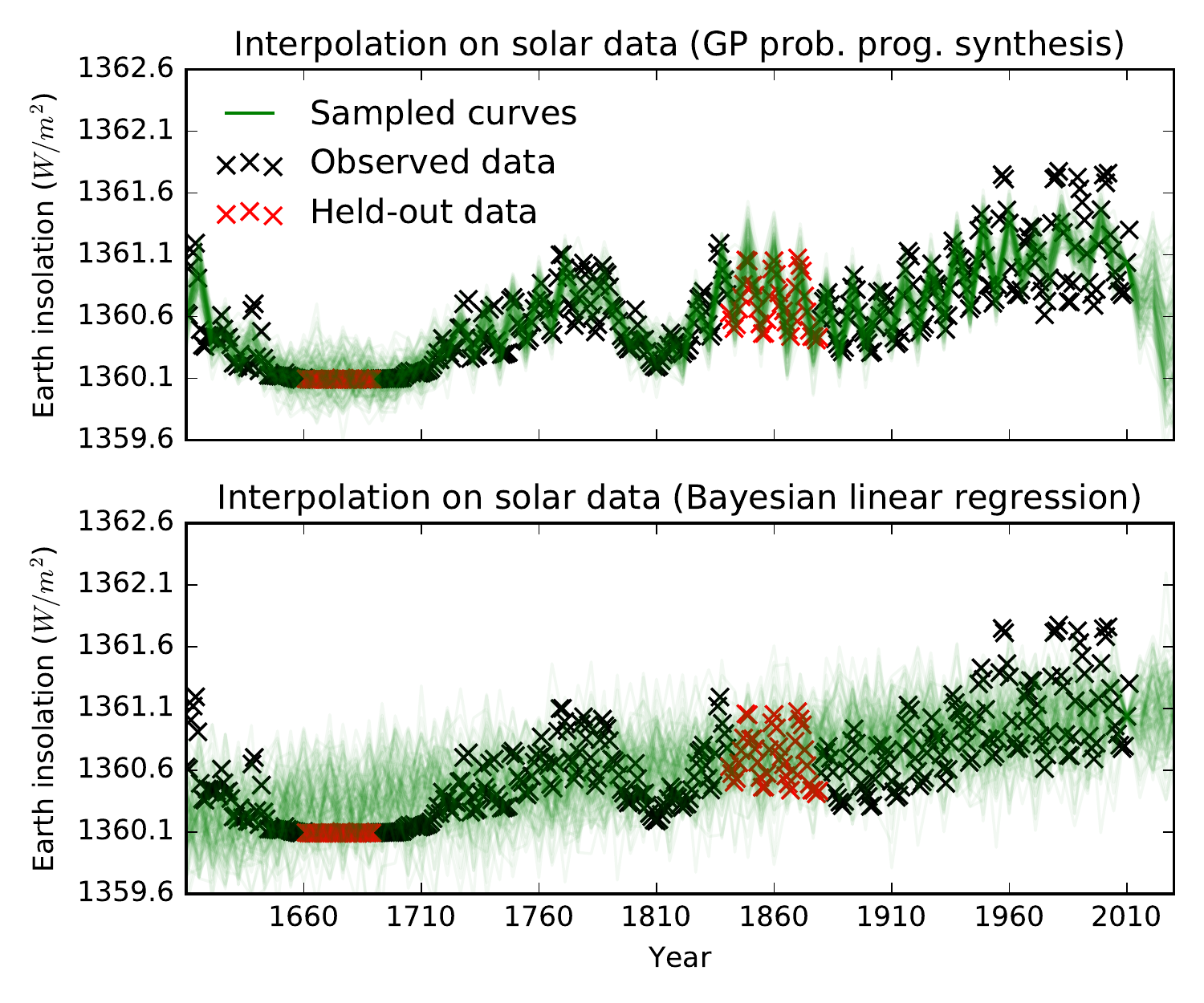}
\captionsetup{skip=0pt}
\subcaption{Interpolation of Earth insolation}
\label{subfig:predictions-interpolation}
\end{subfigure}
\end{subfigure}

\caption{Comparing GP regression via probabilistic program synthesis and
Bayesian linear regression for extrapolating and interpolating on held-out data
(red crosses) in real-world time series. The GP learns combinations of periodic
and change-point structures, improving predictive performance.}
\label{fig:predictions}

\medskip
\hrule

%% Figure: Interpolation and Extrapolation.
\begin{subfigure}{\linewidth}
\includegraphics[width=\textwidth]{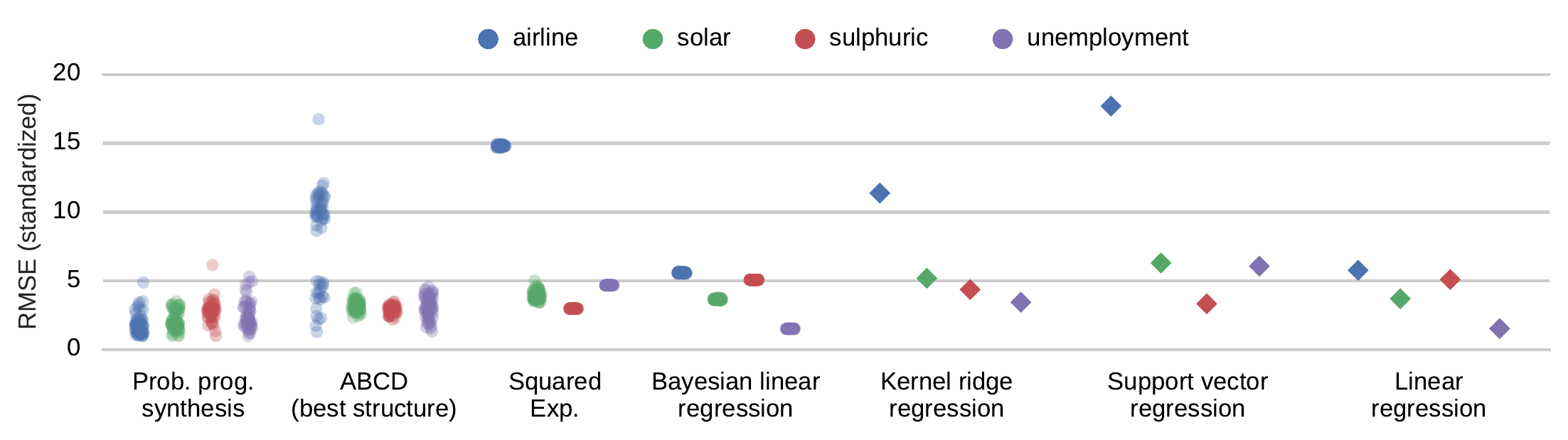}
\end{subfigure}

\caption{Held-out predictive performance of GP regression via probabilistic
program synthesis versus non-parametric regression and standard machine learning
baselines on 4 real-world datasets. Each point for the GP methods is the RMSE of
a posterior sample, standardized to lowest overall error = 1. The ABCD baseline
learns hyperparameters (in Venture) for structures reported by
\cite{lloyd2014}.}
\label{fig:baselines}

\medskip

%% Figure: Lines of Code.

\begin{subtable}{\linewidth}
\centering\footnotesize
\begin{tabular}{lll}
\toprule
  & \textbf{Probabilistic program synthesis}
  & \textbf{ABCD \citep{lloyd2014}}
  \\ \midrule
\textbf{Main system}
  & 69 (Figure~\ref{fig:venturescript})
  & 4,166
  \\
\textbf{Gaussian process libraries}
  & 2,164 (Venture \texttt{gpmem} \citep{schaechtle2015})
  & 13,945 (GPML Toolbox \citep{rasmussen2010})
  \\
\textbf{Generic inference implementation}
  & 1,887 (Venture \citep{mansinghka2014})
  & ---
  \\ \bottomrule
\end{tabular}
\end{subtable}
\caption{Lines of code comparison between probabilistic program synthesis
in Venture, and ABCD.}
\label{tab:lines-of-code}

\end{figure}

We next applied the technique to regression problems on real-world time series.
Figure~\ref{subfig:predictions-extrapolation} shows extrapolation performance on
a dataset of airline passenger volume between 1949 and 1960. The GP detects the
linear trend with periodic variation, leading to very accurate predictions.
Figure~\ref{subfig:predictions-interpolation} shows interpolation on a dataset
of solar radiation between the years 1660 and 2010. The GP successfully models
the qualitative change at around 1760, which correctly results in different
interpolation characteristics at both ends. In contrast, Bayesian linear
regression is forced to treat such structural effects as unmodeled noise.

\FloatBarrier

%!TEX root = ../main.tex

Finally, we compared the predictive performance against six baselines on four
datasets from \citet{lloyd2014}, shown in Figure~\ref{fig:baselines}. The GP
based on probabilistic program synthesis achieved very competitive prediction
error on all tasks.
Figure~\ref{tab:lines-of-code} illustrates that implementing probabilistic
program synthesis in Venture, an expressive probabilistic programming system
with reusable inference machinery, leads to large reductions in code length and
complexity.

\section{Discussion}

We have described and implemented a framework for time series structure
discovery using probabilistic program synthesis. We also have assessed efficacy
of the approach on synthetic and real-world experiments, and demonstrated
improvements in model discovery, extensibility, and predictive accuracy. It
seems promising to apply probabilistic program synthesis to several other
settings, such as fully-Bayesian search in compositional generative grammars for
other model classes \citep{grosse2012}, or Bayes net structure learning with
structured priors \citep{mansinghka2006}. We hope the formalisms in this paper
encourage broader use of probabilistic programming techniques to learn
symbolic structures in other applied domains.

\bibliographystyle{plainnat}
\bibliography{main}

\end{document}